# Modelos Generativos basados en Mecanismos de Difusión

Survey

**Jordi de la Torre**[*]
Ph.D. in Computer Science (ML/AI)
Universitat Oberta de Catalunya
Barcelona, ES
jordi.delatorre@gmail.com

February 18, 2023

## Abstract

Los modelos generativos basados en difusión son un marco de diseño que permite generar nuevas imágenes a partir de procesos análogos a los encontrados en la termodinámica de no equilibrio. Estos modelos modelizan la reversión de un proceso físico de difusión en el que dos líquidos miscibles de diferentes colores se mezclan progresivamente hasta formar una mezcla homogénea. Los modelos de difusión se pueden aplicar a señales de distinta naturaleza, como señales de audio e imagen. Para el caso de imagen, se lleva a cabo un proceso de corrupción progresiva de los píxeles mediante la aplicación de ruido aleatorio, y una red neuronal es entrenada para la reversión de cada uno de los pasos de corrupción. Para que el proceso de reconstrucción resulte reversible, es necesario realizar la corrupción de forma muy progresiva. Si el entrenamiento de la red neuronal es exitoso, será posible generar una imagen a partir de ruido aleatorio encadenando un número de pasos similar al utilizado para la deconstrucción de imágenes en tiempo de entrenamiento. En este artículo presentamos los fundamentos teóricos en los que se basa este método así como algunas de sus aplicaciones. Este artículo está en español para facilitar la llegada de este conocimiento científico a la comunidad hispanohablante.

## Abstract

Diffusion-based generative models are a design framework that allows generating new images from processes analogous to those found in non-equilibrium thermodynamics. These models model the reversal of a physical diffusion process in which two miscible liquids of different colors progressively mix until they form a homogeneous mixture. Diffusion models can be applied to signals of a different nature, such as audio and image signals. In the image case, a progressive pixel corruption process is carried out by applying random noise, and a neural network is trained to revert each one of the corruption steps. For the reconstruction process to be reversible, it is necessary to carry out the corruption very progressively. If the training of the neural network is successful, it will be possible to generate an image from random noise by chaining a number of steps similar to those used for image deconstruction at training time. In this article we present the theoretical foundations on which this method is based as well as some of its applications. This article is in Spanish to facilitate the arrival of this scientific knowledge to the Spanish-speaking community.

[*]mailto:jordi.delatorre@gmail.com web:jorditg.github.io

# 1 Introducción

Los modelos generativos basados en difusión, propuestos por primera vez en [1], son un marco de diseño que permite generar nuevas imágenes mediante la reversión de procesos análogos a los encontrados en la termodinámica de no equilibrio. Si en un vaso de agua introducimos un líquido miscible de un color distinto, observaremos un proceso físico de difusión en el que de los dos líquidos se van mezclando progresivamente hasta finalmente formar una mezcla homogénea. Los modelos basados en difusión modelizan la reversión de este proceso.

Los modelos de difusión se pueden aplicar a señales de distinta naturaleza, entre otras, a señales de audio e imagen. Para el caso de imagen, el proceso se inicia llevando a cabo un proceso de corrupción progresiva de los píxeles que la componen mediante la aplicación de ruido aleatorio, a modo de proceso de difusión natural. Una red neuronal es entrenada para la reversión de cada uno de los mencionados pasos de corrupción. Para que el proceso de reconstrucción resulte reversible es necesario realizar la corrupción de forma muy progresiva. Es habitual que el número de pasos necesarios sea del orden del millar. Si el entrenamiento de dicha red es exitoso será posible generar una imagen a partir de ruido aleatorio encadenando un número de pasos similar al utilizado para la deconstrucción de imágenes en tiempo de entrenamiento.

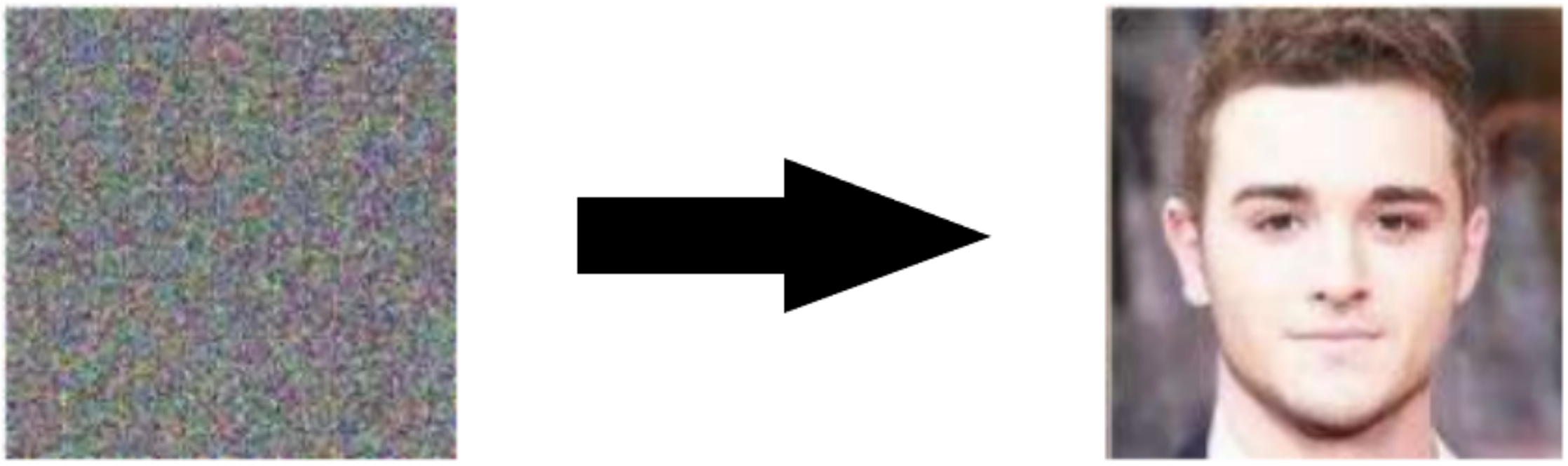

Figura 1: Representación de la generación de imagen típico de un modelo generativo basado en mecanismos de difusión.

# 2 Formulación matemática

Los modelos de difusión son modelos generativos, esto es, son utilizados para generar nuevos datos. El mecanismo de funcionamiento consiste en dos etapas: una primera donde se procede a la corrupción incremental de los datos de entrada mediante de adición de ruido Gaussiano y una segunda donde se aprende a revertir cada uno de los estadios de corrupción aplicados previamente. Después del entrenamiento, las redes de reconstrucción entrenadas son capaces de generar nuevos datos a partir de ruido gaussiano mediante la aplicación sucesiva del proceso de reconstrucción previamente entrenado.

Una cadena de Markov es un proceso estocástico en el cual la probabilidad de que suceda un evento depende únicamente del estado precedente, esto es, toda la información necesaria para predecir el estado siguiente está contenida en el estado actual, siendo, por tanto, independiente de todos los anteriores.

Un modelo de difusión puede ser modelizado como una cadena de Markov donde los nodos representan los sucesivos estadios de la reconstrucción de los datos desde su estadio original hasta el final, representativo de los datos reconstruidos.

Para revertir el proceso debemos ser capaces de modelizar $q(\boldsymbol{x}_t|\boldsymbol{x}_{t-1})$. La figura 1 representa el proceso al que se somete cada imagen durante el entrenamiento. La cadena de Markov del proceso de difusión directa realiza una deconstrucción paulatina de los datos, añadiendo ruido de forma gradual, para obtener una aproximación al posterior $q(\boldsymbol{x}_{1:T}|\boldsymbol{x}_0)$, donde $\boldsymbol{x}_1, ..., \boldsymbol{x}_T$, son variables latentes de las misma dimensiones que $\boldsymbol{x}_0$. Al final del proceso se acaba transformando la imagen original en ruido gaussiano.

La red generativa utiliza la información generada como medio de optimización del modelo de reconstrucción del proceso inverso. La función generativa corresponde a la distribución de probabilidad $p_\theta(\boldsymbol{x}_{t-1}|\boldsymbol{x}_t)$. La cadena de Markov representativa del proceso de reconstrucción se muestra en la figura 2.

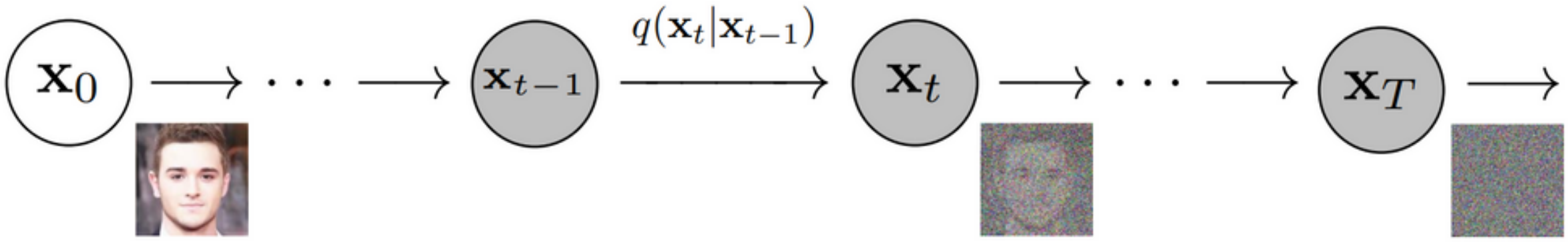

Figura 2: Representación de la cadena de Markov representativa del proceso de deconstrucción de una imagen.

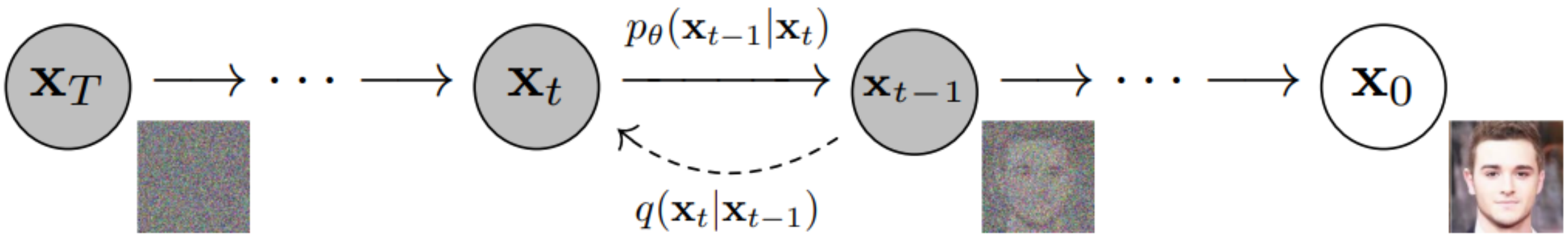

Figura 3: Representación de la cadena de Markov representativa del proceso de reconstrucción de una imagen.

Los modelos basados en difusión son capaces de generar imágenes de alta calidad sin requerir entrenamiento adversario típico de las GANs. Las dificultades de la optimización de las redes GAN ya ha sido tratada en el capítulo correspondiente. Los modelos de difusión presentan beneficios de escalabilidad y paralelización respecto a otros tipos de modelos generativos como el mencionado.

## 3 Modelos probabilísticos de eliminación del ruido producido por difusión

Los modelos de difusión son modelos de variables latentes de la forma $p_\theta(\boldsymbol{x}_0) = \int p_\theta(\boldsymbol{x}_0, ..., \boldsymbol{x}_T)d\boldsymbol{x}_1...d\boldsymbol{x}_T$, donde $\boldsymbol{x}_1...\boldsymbol{x}_T$ son variables latentes de la mismas dimensiones que los datos $\boldsymbol{x}_0 \sim q(\boldsymbol{x}_0)$.

### 3.1 Proceso directo de difusión

Dado un punto de muestreo $\boldsymbol{x}_0 \sim q(\boldsymbol{x})$, definimos el proceso directo de difusión como aquel en el que añadimos ruido Gaussiano a la muestra original en $T$ pasos sucesivos, produciendo una secuencia de muestras $\boldsymbol{x}_1, ..., \boldsymbol{x}_T$. El tamaño del cambio se controla mediante el parámetro $\{\beta_t \in (0,1)\}_{t=1}^{T}$.

$$q(\boldsymbol{x}_t|\boldsymbol{x}_{t-1}) = \mathcal{N}(\boldsymbol{x}_t; \sqrt{1-\beta_t}\boldsymbol{x}_{t-1}, \beta_t\boldsymbol{I})$$
$$q(\boldsymbol{x}_1, ..., \boldsymbol{x}_T|\boldsymbol{x}_0) = \prod_{t=1}^{T} q(\boldsymbol{x}_t|\boldsymbol{x}_{t-1})$$

La muestra $\boldsymbol{x}_0$ pierde gradualmente sus atributos distinguibles. Con $T \to \infty$, $\boldsymbol{x}_T$ tiende a una distribución isotrópica Gaussiana, p.e. $\mathcal{N}(\boldsymbol{0}, \boldsymbol{I})$.

Una propiedad interesante que a continuación demostramos, es que para cualquier $t$ es posible obtener en un único paso la muestra $\boldsymbol{x}_t$:

Sea $\alpha_t = 1 - \beta_t$ y $\overline{\alpha}_t = \prod_{i=1}^{t} \alpha_i$ entonces,

$$\begin{aligned}
\boldsymbol{x}_t &= \sqrt{\alpha_t}\boldsymbol{x}_{t-1} + \sqrt{1-\alpha_t}\boldsymbol{\epsilon}_{t-1}; && \boldsymbol{\epsilon}_i \sim \mathcal{N}(\boldsymbol{0}, \boldsymbol{1}) \\
&= \sqrt{\alpha_t\alpha_{t-1}}\boldsymbol{x}_{t-2} + \sqrt{1-\alpha_t\alpha_{t-1}}\overline{\boldsymbol{\epsilon}}_{t-2}; && \overline{\boldsymbol{\epsilon}}_{t-2} \sim \mathcal{N}(\boldsymbol{0}, \sqrt{1-\alpha_t\alpha_{t-1}}) \\
&= ... \\
&= \sqrt{\overline{\alpha}_t}\boldsymbol{x}_0 + \sqrt{1-\overline{\alpha}_t}\boldsymbol{\epsilon} \\
q(\boldsymbol{x}_t|\boldsymbol{x}_0) &= \mathcal{N}(\boldsymbol{x}_t; \sqrt{\overline{\alpha}_t}\boldsymbol{x}_0, (1-\overline{\alpha}_t)\boldsymbol{I})
\end{aligned}$$

Por lo general, conforme se avanza en el proceso de corrupción, es posible realizar pasos de actualización más grandes, por lo que $\beta_1 < \beta_2 < ... < \beta_T$ y, por tanto, $\overline{\alpha}_1 > \overline{\alpha}_2 > ... > \overline{\alpha}_T$.

Hasta ahora hemos visto cómo derivar $q(\boldsymbol{x}_t, \boldsymbol{x}_0)$. Para derivar $q(\boldsymbol{x}_t)$ sabemos que $q(\boldsymbol{x}_t) = \int q(\boldsymbol{x}_0, \boldsymbol{x}_t) d\boldsymbol{x}_0 = \int q(\boldsymbol{x}_t|\boldsymbol{x}_0) q(\boldsymbol{x}_0) d\boldsymbol{x}_0$, entonces podemos muestrear $\boldsymbol{x}_t \sim q(\boldsymbol{x}_t)$ en dos pasos: primero muestreando $\boldsymbol{x}_0 \sim q(\boldsymbol{x}_0)$ de los datos de entrada y después muestreando $\boldsymbol{x}_t \sim q(\boldsymbol{x}_t|\boldsymbol{x}_0)$, esto es, muestreo ancestral.

A continuación vamos a derivar la expresión de $q(\boldsymbol{x}_{t-1}|\boldsymbol{x}_t, \boldsymbol{x}_0)$. Combinando la aplicación de la regla de bayes y la expresión de la función gaussiana tenemos:

$$\begin{aligned} q(\mathbf{x}_{t-1}|\mathbf{x}_t, \mathbf{x}_0) &= q(\mathbf{x}_t|\mathbf{x}_{t-1}, \mathbf{x}_0) \frac{q(\mathbf{x}_{t-1}|\mathbf{x}_0)}{q(\mathbf{x}_t|\mathbf{x}_0)} \\ &\propto \exp\Big( -\frac{1}{2}\big(\frac{(\mathbf{x}_t - \sqrt{\alpha_t}\mathbf{x}_{t-1})^2}{\beta_t} + \frac{(\mathbf{x}_{t-1} - \sqrt{\bar{\alpha}_{t-1}}\mathbf{x}_0)^2}{1-\bar{\alpha}_{t-1}} - \frac{(\mathbf{x}_t - \sqrt{\bar{\alpha}_t}\mathbf{x}_0)^2}{1-\bar{\alpha}_t}\big)\Big) \\ &= \exp\Big( -\frac{1}{2}\big(\frac{\mathbf{x}_t^2 - 2\sqrt{\alpha_t}\mathbf{x}_t\mathbf{x}_{t-1} + \alpha_t\mathbf{x}_{t-1}^2}{\beta_t} + \frac{\mathbf{x}_{t-1}^2 - 2\sqrt{\bar{\alpha}_{t-1}}\mathbf{x}_0\mathbf{x}_{t-1} + \bar{\alpha}_{t-1}\mathbf{x}_0^2}{1-\bar{\alpha}_{t-1}} - \frac{(\mathbf{x}_t - \sqrt{\bar{\alpha}_t}\mathbf{x}_0)^2}{1-\bar{\alpha}_t}\big)\Big) \\ &= \exp\Big( -\frac{1}{2}\big((\frac{\alpha_t}{\beta_t} + \frac{1}{1-\bar{\alpha}_{t-1}})\mathbf{x}_{t-1}^2 - (\frac{2\sqrt{\alpha_t}}{\beta_t}\mathbf{x}_t + \frac{2\sqrt{\bar{\alpha}_{t-1}}}{1-\bar{\alpha}_{t-1}}\mathbf{x}_0)\mathbf{x}_{t-1} + C(\mathbf{x}_t, \mathbf{x}_0)\big)\Big) \end{aligned}$$

Expresando lo términos en función de una distribución estándar gaussiana tenemos que:

$$\begin{aligned} \tilde{\beta}_t &= 1/(\frac{\alpha_t}{\beta_t} + \frac{1}{1-\bar{\alpha}_{t-1}}) = 1/(\frac{\alpha_t - \bar{\alpha}_t + \beta_t}{\beta_t(1-\bar{\alpha}_{t-1})}) = \frac{1-\bar{\alpha}_{t-1}}{1-\bar{\alpha}_t} \cdot \beta_t \\ \tilde{\boldsymbol{\mu}}_t(\mathbf{x}_t, \mathbf{x}_0) &= (\frac{\sqrt{\alpha_t}}{\beta_t}\mathbf{x}_t + \frac{\sqrt{\bar{\alpha}_{t-1}}}{1-\bar{\alpha}_{t-1}}\mathbf{x}_0)/(\frac{\alpha_t}{\beta_t} + \frac{1}{1-\bar{\alpha}_{t-1}}) \\ &= (\frac{\sqrt{\alpha_t}}{\beta_t}\mathbf{x}_t + \frac{\sqrt{\bar{\alpha}_{t-1}}}{1-\bar{\alpha}_{t-1}}\mathbf{x}_0)\frac{1-\bar{\alpha}_{t-1}}{1-\bar{\alpha}_t} \cdot \beta_t \\ &= \frac{\sqrt{\alpha_t}(1-\bar{\alpha}_{t-1})}{1-\bar{\alpha}_t}\mathbf{x}_t + \frac{\sqrt{\bar{\alpha}_{t-1}}\beta_t}{1-\bar{\alpha}_t}\mathbf{x}_0 \end{aligned}$$

Sabemos que $\mathbf{x}_0 = \frac{1}{\sqrt{\bar{\alpha}_t}}(\mathbf{x}_t - \sqrt{1-\bar{\alpha}_t}\boldsymbol{\epsilon}_t)$, introduciendo esta expresión en la de $\tilde{\mu}_t$ obtenemos:

$$\begin{aligned} \tilde{\boldsymbol{\mu}}_t(\boldsymbol{x}_t, \boldsymbol{\epsilon}_t) &= \frac{\sqrt{\alpha_t}(1-\bar{\alpha}_{t-1})}{1-\bar{\alpha}_t}\mathbf{x}_t + \frac{\sqrt{\bar{\alpha}_{t-1}}\beta_t}{1-\bar{\alpha}_t}\frac{1}{\sqrt{\bar{\alpha}_t}}(\mathbf{x}_t - \sqrt{1-\bar{\alpha}_t}\boldsymbol{\epsilon}_t) \\ &= \frac{1}{\sqrt{\alpha_t}}\Big(\mathbf{x}_t - \frac{1-\alpha_t}{\sqrt{1-\bar{\alpha}_t}}\boldsymbol{\epsilon}_t\Big) \end{aligned}$$

A pesar de que $q(\boldsymbol{x}_{t-1}|\boldsymbol{x}_t)$ es intratable, se demuestra que $q(\boldsymbol{x}_{t-1}|\boldsymbol{x}_t, \boldsymbol{x}_0)$ es tratable, $q(\mathbf{x}_{t-1}|\mathbf{x}_t, \mathbf{x}_0) = \mathcal{N}(\mathbf{x}_{t-1}; \tilde{\boldsymbol{\mu}}(\mathbf{x}_t, \mathbf{x}_0), \tilde{\beta}_t\mathbf{I})$.

### 3.2 Proceso de difusión inverso o de reconstrucción

Si fuera posible revertir el proceso descrito en el apartado anterior y obtener muestras de la distribución $q(\boldsymbol{x}_{t-1}|\boldsymbol{x}_t)$ entonces sería posible reconstruir una muestra real a partir de una entrada de ruido Gaussiano, $\boldsymbol{x}_T \sim \mathcal{N}(\mathbf{0}, \boldsymbol{I})$.

Una forma de hacerlo es estimar $q(\boldsymbol{x}_{t-1}|\boldsymbol{x}_t)$ mediante el uso de un modelo paramétrico $p_\theta$ que aproxime dichas probabilidades condicionales. Para $\beta_t$ suficientemente pequeño, $q(\boldsymbol{x}_{t-1}|\boldsymbol{x}_t)$ se puede aproximar a una distribución normal. Se decide aproximar $q(\boldsymbol{x}_{t-1}|\boldsymbol{x}_t)$ mediante un modelo paramétrico $p_\theta$ como una distribución gaussiana y parametrizar su media y su varianza:

$$\begin{gathered} p(\boldsymbol{x}_T) = \mathcal{N}(\boldsymbol{x}_T; \mathbf{0}, \boldsymbol{I}) \\ p_\theta(\boldsymbol{x}_{t-1}|\boldsymbol{x}_t) = \mathcal{N}(\boldsymbol{x}_{t-1}; \boldsymbol{\mu}_\theta(\boldsymbol{x}_t, t), \boldsymbol{\Sigma}_\theta(\boldsymbol{x}_t, t))) \\ \boldsymbol{\Sigma}_\theta(\boldsymbol{x}_t, t) = \sigma_t\boldsymbol{I} \end{gathered}$$

$\mu_\theta(\boldsymbol{x}_t, t)$ se suele parametrizar como una red neuronal (comúnmente utilizando la arquitectura U-Net o bien el auto-codificador atenuador de ruido).

Si aplicamos la ecuación de inversión para todos los pasos (trayectoria) podemos ir de $\boldsymbol{x}_T$ hasta la distribución de los datos:

$$p_\theta(\boldsymbol{x}_0, ..., \boldsymbol{x}_T) = p_\theta(\boldsymbol{x}_T)\prod_{t=1}^{T} p_\theta(\boldsymbol{x}_{t-1}|\boldsymbol{x}_t)$$

### 3.3 Función de pérdida

Para poder encontrar los parámetros óptimos del modelo de inversión, debemos buscar una forma de conectar el proceso de difusión directa (representado por la distribución $q$) con el proceso inverso (distribución $p_\theta$). Una función objetivo podría ser la siguiente:

$$-\log p(\boldsymbol{x}_0) \leq -\log p(\boldsymbol{x}_0) + D_{KL}(q(\boldsymbol{x}_1, ..., \boldsymbol{x}_T|\boldsymbol{x}_0)||p_\theta(\boldsymbol{x}_1, ..., \boldsymbol{x}_T|\boldsymbol{x}_0))$$

En el caso ideal que $q(\boldsymbol{x}_1, ..., \boldsymbol{x}_T|\boldsymbol{x}_0)$ y $p_\theta(\boldsymbol{x}_1, ..., \boldsymbol{x}_T|\boldsymbol{x}_0)$ fueran iguales, la desigualdad se convertiría en igualdad.

Manipulando la expresión anterior:

$$\begin{aligned}
-\log p(\boldsymbol{x}_0) &\leq -\log p(\boldsymbol{x}_0) + D_{KL}(q(\boldsymbol{x}_1, ..., \boldsymbol{x}_T|\boldsymbol{x}_0)||p_\theta(\boldsymbol{x}_1, ..., \boldsymbol{x}_T|\boldsymbol{x}_0)) \\
&= -\log p(\boldsymbol{x}_0) + \mathbb{E}_{\boldsymbol{x}_1,...,\boldsymbol{x}_T} \sim q(\boldsymbol{x}_1, ..., \boldsymbol{x}_T|\boldsymbol{x}_\mathbf{0}) \Big[ \log \frac{q(\boldsymbol{x}_1,...,\boldsymbol{x}_T|\boldsymbol{x}_0)}{p_\theta(\boldsymbol{x}_0,...,\boldsymbol{x}_T)/p_\theta(\boldsymbol{x}_0)} \Big] \\
&= -\log p(\boldsymbol{x}_0) + \mathbb{E}_q \Big[ \log \frac{q(\boldsymbol{x}_1,...,\boldsymbol{x}_T|\boldsymbol{x}_0)}{p_\theta(\boldsymbol{x}_0,...,\boldsymbol{x}_T)} + \log p_\theta(\boldsymbol{x}_0) \Big] \\
&= \mathbb{E}_q \Big[ \log \frac{q(\boldsymbol{x}_1,...,\boldsymbol{x}_T|\boldsymbol{x}_0)}{p_\theta(\boldsymbol{x}_0,...,\boldsymbol{x}_T)} \Big] = L_{VLB} \\
L_{VLB} &\geq -\mathbb{E}_{q(\boldsymbol{x}_0)} \log p_\theta(\boldsymbol{x}_\mathbf{0})
\end{aligned}$$

Se puede llegar al mismo resultado partiendo de la desigualdad de Jensen:

$$\begin{aligned}
L_{CE} &= -\mathbb{E}_{q(\boldsymbol{x}_0)} \log p_\theta(\boldsymbol{x}_\mathbf{0}) \\
&= -\mathbb{E}_{q(\boldsymbol{x}_0)} \log \Big( \int p_\theta(\boldsymbol{x}_0, ..., \boldsymbol{x}_T) d\boldsymbol{x}_1...d\boldsymbol{x}_T \Big) \\
&= -\mathbb{E}_{q(\boldsymbol{x}_0)} \log \Big( \int q(\boldsymbol{x}_1, ..., \boldsymbol{x}_T|\boldsymbol{x}_0) \frac{p_\theta(\boldsymbol{x}_0,...,\boldsymbol{x}_T)}{q(\boldsymbol{x}_1,...,\boldsymbol{x}_T|\boldsymbol{x}_0)} d\boldsymbol{x}_1...d\boldsymbol{x}_T \Big) \\
&= \mathbb{E}_{q(\boldsymbol{x}_0)} \log \Big( \mathbb{E}_{q(\boldsymbol{x}_1,...,\boldsymbol{x}_T|\boldsymbol{x}_0)} \frac{p_\theta(\boldsymbol{x}_0,...,\boldsymbol{x}_T)}{q(\boldsymbol{x}_1,...,\boldsymbol{x}_T|\boldsymbol{x}_0)} \Big) \\
&\leq -\mathbb{E}_{q(\boldsymbol{x}_0,...,\boldsymbol{x}_T)} \log \Big( \frac{p_\theta(\boldsymbol{x}_0,...,\boldsymbol{x}_T)}{q(\boldsymbol{x}_1,...,\boldsymbol{x}_T|\boldsymbol{x}_0)} \Big) \\
&= \mathbb{E}_{q(\boldsymbol{x}_0,...,\boldsymbol{x}_T)} \log \Big( \frac{q(\boldsymbol{x}_1,...,\boldsymbol{x}_T|\boldsymbol{x}_0)}{p_\theta(\boldsymbol{x}_0,...,\boldsymbol{x}_T)} \Big) = L_{VLB}
\end{aligned}$$

Para convertir cada término de la ecuación para que sea analíticamente computable, el objetivo se puede reescribir como una combinación de varios términos de divergencia KL y entropía (Apéndice B de [1]) llegando a la siguiente expresión:

$$\begin{aligned}
L_{VLB} &= \mathbb{E}_{q(\boldsymbol{x}_0,...,\boldsymbol{x}_T)} \Big[ \log \frac{q(\boldsymbol{x}_1,...,\boldsymbol{x}_T|\boldsymbol{x}_0)}{p_\theta(\boldsymbol{x}_0,...,\boldsymbol{x}_T)} \Big] \\
&= \mathbb{E}_q \Big[ \log \frac{\prod_{t=1}^T q(\boldsymbol{x}_t|\boldsymbol{x}_{t-1})}{p_\theta(\boldsymbol{x}_T) \prod_{t=1}^T p_\theta(\boldsymbol{x}_{t-1}|\boldsymbol{x}_t)} \Big] \\
&= \mathbb{E}_q \Big[ -\log p_\theta(\boldsymbol{x}_T) + \sum_{t=1}^T \log \frac{q(\boldsymbol{x}_t|\boldsymbol{x}_{t-1})}{p_\theta(\boldsymbol{x}_{t-1}|\boldsymbol{x}_t)} \Big] \\
&= \mathbb{E}_q \Big[ -\log p_\theta(\boldsymbol{x}_T) + \sum_{t=2}^T \log \frac{q(\boldsymbol{x}_t|\boldsymbol{x}_{t-1})}{p_\theta(\boldsymbol{x}_{t-1}|\boldsymbol{x}_t)} + \log \frac{q(\boldsymbol{x}_1|\boldsymbol{x}_0)}{p_\theta(\boldsymbol{x}_0|\boldsymbol{x}_1)} \Big] \\
&= \mathbb{E}_q \Big[ -\log p_\theta(\boldsymbol{x}_T) + \sum_{t=2}^T \log \Big( \frac{q(\boldsymbol{x}_{t-1}|\boldsymbol{x}_t,\boldsymbol{x}_0)}{p_\theta(\boldsymbol{x}_{t-1}|\boldsymbol{x}_t)} \cdot \frac{q(\boldsymbol{x}_t|\boldsymbol{x}_0)}{q(\boldsymbol{x}_{t-1}|\boldsymbol{x}_0)} \Big) + \log \frac{q(\boldsymbol{x}_1|\boldsymbol{x}_0)}{p_\theta(\boldsymbol{x}_0|\boldsymbol{x}_1)} \Big] \\
&= \mathbb{E}_q \Big[ -\log p_\theta(\boldsymbol{x}_T) + \sum_{t=2}^T \log \frac{q(\boldsymbol{x}_{t-1}|\boldsymbol{x}_t,\boldsymbol{x}_0)}{p_\theta(\boldsymbol{x}_{t-1}|\boldsymbol{x}_t)} + \sum_{t=2}^T \log \frac{q(\boldsymbol{x}_t|\boldsymbol{x}_0)}{q(\boldsymbol{x}_{t-1}|\boldsymbol{x}_0)} + \log \frac{q(\boldsymbol{x}_1|\boldsymbol{x}_0)}{p_\theta(\boldsymbol{x}_0|\boldsymbol{x}_1)} \Big] \\
&= \mathbb{E}_q \Big[ -\log p_\theta(\boldsymbol{x}_T) + \sum_{t=2}^T \log \frac{q(\boldsymbol{x}_{t-1}|\boldsymbol{x}_t,\boldsymbol{x}_0)}{p_\theta(\boldsymbol{x}_{t-1}|\boldsymbol{x}_t)} + \log \frac{q(\boldsymbol{x}_T|\boldsymbol{x}_0)}{q(\boldsymbol{x}_1|\boldsymbol{x}_0)} + \log \frac{q(\boldsymbol{x}_1|\boldsymbol{x}_0)}{p_\theta(\boldsymbol{x}_0|\boldsymbol{x}_1)} \Big] \\
&= \mathbb{E}_q \Big[ \log \frac{q(\boldsymbol{x}_T|\boldsymbol{x}_0)}{p_\theta(\boldsymbol{x}_T)} + \sum_{t=2}^T \log \frac{q(\boldsymbol{x}_{t-1}|\boldsymbol{x}_t,\boldsymbol{x}_0)}{p_\theta(\boldsymbol{x}_{t-1}|\boldsymbol{x}_t)} - \log p_\theta(\boldsymbol{x}_0|\boldsymbol{x}_1) \Big] \\
&= \mathbb{E}_q [\underbrace{D_{KL}(q(\boldsymbol{x}_T|\boldsymbol{x}_0) \parallel p_\theta(\boldsymbol{x}_T))}_{L_T} + \sum_{t=2}^T \underbrace{D_{KL}(q(\boldsymbol{x}_{t-1}|\boldsymbol{x}_t, \boldsymbol{x}_0) \parallel p_\theta(\boldsymbol{x}_{t-1}|\boldsymbol{x}_t))}_{L_{t-1}} \underbrace{- \log p_\theta(\boldsymbol{x}_0|\boldsymbol{x}_1)}_{L_0}]
\end{aligned}$$

donde $q(\boldsymbol{x}_{t-1}|\boldsymbol{x}_t, \boldsymbol{x}_0)$ es la versión tratable de la distribución del posterior.

Etiquetando cada componente del ELBO separadamente tenemos:

$$
\begin{aligned}
&L_{VLB} = L_T + L_{T-1} + \cdots + L_0 \\
\text{donde} \\
&L_T = D_{KL}(q(\boldsymbol{x}_T|\boldsymbol{x}_0) \parallel p_\theta(\boldsymbol{x}_T)) \\
&L_t = D_{KL}(q(\boldsymbol{x}_t|\boldsymbol{x}_{t+1}, \boldsymbol{x}_0) \parallel p_\theta(\boldsymbol{x}_t|\boldsymbol{x}_{t+1})) \text{ para } 1 \leq t \leq T-1 \\
&L_0 = -\log p_\theta(\boldsymbol{x}_0|\boldsymbol{x}_1)
\end{aligned}
$$

Analicemos cada uno de los términos:

1. $\mathbb{E}_{q(x_1|x_0)}[\log p_\theta(\boldsymbol{x}_0|\boldsymbol{x}_1)]$ Puede ser interpretado como un término de reconstrucción similar al ELBO de un autocodificador variacional. En [2] modeliza $L_0$ utilizando un decodificador discreto separado derivado de $\mathcal{N}(\boldsymbol{x}_o; \boldsymbol{\mu}_\theta(\boldsymbol{x}_1, 1), \boldsymbol{\Sigma}_\theta(\boldsymbol{x}_1, 1))$
2. $D_{KL}(q(\boldsymbol{x}_T|\boldsymbol{x}_0) \parallel p_\theta(\boldsymbol{x}_T))$ muestra cuánto de cerca se encuentra $\boldsymbol{x}_T$ de la función Gaussiana estándard. El término no tiene parámetros entrenables y, por tanto, es ignorado durante el entrenamiento.
3. $D_{KL}(q(\boldsymbol{x}_t|\boldsymbol{x}_{t+1}, \boldsymbol{x}_0) \parallel p_\theta(\boldsymbol{x}_t|\boldsymbol{x}_{t+1}))$ para $1 \leq t \leq T-1$, calcula la diferencia entre cada uno de los pasos inversos $p_\theta(\boldsymbol{x}_{t-1}|\boldsymbol{x}_t)$ y los aproximados $q(\boldsymbol{x}_{t-1}|\boldsymbol{x}_t, \boldsymbol{x}_0)$

Todos los términos de $L_{VLB}$ (excepto $L_0$) establecen comparaciones entre distribuciones Gaussianas y, por tanto, pueden ser calculadas de forma exacta. A través del ELBO, maximizar la probabilidad se reduce a aprender los pasos de eliminación del ruido $L_t$.

Como tanto $q(\boldsymbol{x}_{t-1}|\boldsymbol{x}_t)$ como $p_\theta(\boldsymbol{x}_{t-1}|\boldsymbol{x}_t)$ son distribuciones normales, la divergencia de Kullback-Leibler se puede expresar de forma analítica como:

$$
L_{t-1} = D_{KL}(q(\boldsymbol{x}_{t-1}|\boldsymbol{x}_t, \boldsymbol{x}_0)||p_\theta(\boldsymbol{x}_{t-1}|\boldsymbol{x}_t)) = \mathbb{E}_q\left[\frac{1}{2\sigma_t^2}||\tilde{\mu}_t(\boldsymbol{x}_t, \boldsymbol{x}_0) - \mu_\theta(\boldsymbol{x}_t, t)||^2\right] + C
$$

Sabemos que $\boldsymbol{x}_t = \sqrt{\overline{\alpha}_t}\boldsymbol{x}_0 + \sqrt{1-\overline{\alpha}_t}$ y que $\tilde{\mu}_t(\boldsymbol{x}_t, \boldsymbol{x}_0) = \frac{1}{\sqrt{1-\beta_t}}\left(\boldsymbol{x}_t - \frac{\beta_t}{\sqrt{1-\overline{\alpha}_t}}\right)$, entonces se puede representar la función paramétrica $\mu_\theta(\boldsymbol{x}_t, t)$ como un modelo paramétrico de predicción del ruido $\epsilon_\theta(\boldsymbol{x}_t, t)$:

$$
\boldsymbol{\mu}_\theta(\boldsymbol{x}_t, t) = \frac{1}{\sqrt{\boldsymbol{\alpha}_t}}\left(\boldsymbol{x}_t - \frac{\boldsymbol{\beta}_t}{\sqrt{1-\overline{\boldsymbol{\alpha}}_t}}\epsilon_\theta(\boldsymbol{x}_t, t)\right)
$$

La función de pérdida quedaría representada de la forma siguiente:

$$
\begin{aligned}
L_{t-1} &= \mathbb{E}_{\boldsymbol{x}_0 \sim q(\boldsymbol{x}_0), \epsilon \sim \mathcal{N}(\boldsymbol{0}, \boldsymbol{I})}\left[\lambda_t||\epsilon - \epsilon_\theta(\boldsymbol{x}_t, t)||^2\right] + C \\
\lambda_t &= \frac{\beta_t^2}{2\sigma_t^2(1-\beta_t)(1-\overline{\alpha}_t)} \\
\boldsymbol{x}_t &= \sqrt{\overline{\alpha}_t}\boldsymbol{x}_0 + \sqrt{1-\overline{\alpha}_t}\epsilon
\end{aligned}
$$

El valor de $\lambda_t$ es un peso que pondera la importancia en función del tiempo. Este valor suele ser bastante grande para valores de $t$ pequeños. En [2] observan que estableciendo $\lambda_t = 1$ mejora la calidad del muestreo. Alineado con esta decisión ellos usan la función de pérdida siguiente:

$$
L_{\text{simple}} = \mathbb{E}_{x_0 \sim q(x_0), \epsilon \sim \mathcal{N}(\boldsymbol{0}, \boldsymbol{I}), l \sim \mathcal{U}(1, T)}\left[||\epsilon - \epsilon_\theta(\boldsymbol{x}_t, t)||^2\right]
$$

Otras estrategias avanzadas de ponderación de $\lambda_t$ pueden consultarse en [3].

### 3.4 Proceso de muestreo y entrenamiento

Los algoritmos 1 y 2 presentados a continuación describen el proceso de entrenamiento y de muestreo respectivamente presentado en esta sección.

**Algorithm 1** Difusión: Algoritmo de entrenamiento

**1)** **repetir**
**2)** $\boldsymbol{x}_0 \sim q(\boldsymbol{x}_0)$
**3)** $t \sim \text{Uniform}(\{1, ..., T\})$
**4)** $\boldsymbol{\epsilon} \sim \mathcal{N}(\mathbf{0}, \boldsymbol{I})$
**5)** Descenso de gradiente con $\nabla_\theta \|\boldsymbol{\epsilon}_t - \boldsymbol{\epsilon}_\theta(\sqrt{\bar{\alpha}_t}\mathbf{x}_0 + \sqrt{1-\bar{\alpha}_t}\boldsymbol{\epsilon}_t, t)\|^2$
**6)** **hasta** convergencia

**Algorithm 2** Difusión: Algoritmo de muestreo

**1)** $\boldsymbol{x}_T \sim \mathcal{N}(\mathbf{0}, \boldsymbol{I})$
**2)** **para** t = T,..., 1 **hacer**
**3)** $\boldsymbol{z} \sim \mathcal{N}(\mathbf{0}, \boldsymbol{I})$ si $t > 1$ sino $\boldsymbol{z} = \mathbf{0}$
**4)** $\boldsymbol{x}_{t-1} = \frac{1}{\sqrt{\boldsymbol{\alpha}_t}}\left(\boldsymbol{x}_t - \frac{1-\boldsymbol{\alpha}_t}{\sqrt{1-\bar{\boldsymbol{\alpha}}_t}}\epsilon_\theta(\boldsymbol{x}_t, t)\right) + \boldsymbol{\sigma}_t\boldsymbol{z}$
**5)** **fpara**
**6)** **retorna** $\boldsymbol{x}_0$

### 3.5 Implementación

Los modelos basados en mecanismos de difusión presentados hasta el momento para representar $\epsilon_\theta(\boldsymbol{x}_t, t)$ varían ligeramente dependiendo de la publicación pero suelen coincidir en la utilización de arquitecturas basadas en U-Net [4] con bloques Wide ResNet [5] y etapas de autoatención. La representación de $t$ se suele hacer o bien mediante los embeddings posicionales típicos de los transformadores [6] o bien mediante atributos de Fourier aleatorios [7]. No hay ninguna razón que impida que futuras arquitecturas utilicen otras propuestas distintas.

### 3.6 Parametrización de $\beta_t$ y $\Sigma_\theta$

Hemos visto que $\beta_t$ y $\sigma_t^2$ controlan la varianza del proceso de difusión directo e inverso respectivamente. Es habitual utilizar un sistema de distribución lineal para $\beta_t$ y fijar $\Sigma_\theta = \sigma_t^2 = \beta_t$.

Por ejemplo, [2] las varianzas $\beta_t$ se fijan de modo que sean una secuencia de constantes crecientes desde $\beta_1 = 10^{-4}$ hasta $\beta_T = 0{,}02$. Son relativamente pequeñas comparadas con el rango normalizado de los valores admisibles para los píxeles, que en dicho trabajo está en el intervalo $[-1, 1]$. En dicho trabajo. la varianza del proceso inverso no es un valor paramétrico sino que es una constante $\boldsymbol{\Sigma}_\theta(\boldsymbol{x}_t, t) = \sigma_t\boldsymbol{I}$, donde $\sigma_t$ es $\beta_t$ o bien $\tilde{\beta}_t = \frac{1-\bar{\alpha}_{t-1}}{1-\bar{\alpha}_t}\beta_t$. En dicho artículo las muestras generadas son de alta calidad pero aún no consiguen valores competitivos en lo referente a verosimilitud.

En [8] se proponen varias técnicas para conseguir mejores resultados en lo que a verosimilitud se refiere. Una de dichas mejoras consiste cambiar la evolución lineal de $\beta_t$ por una tipo coseno. El objetivo del cambio no es tanto la función escogida en sí, sino el hecho de conseguir un cambio pequeño en los dos extremos $1$ y $T$ y un cambio lineal en los valores intermedios. Para $\boldsymbol{\Sigma}_\theta(\boldsymbol{x}_t, t)$ proponen interpolar entre los valores $\beta_t$ y $\tilde{\beta}_t$ mediante la función $\boldsymbol{\Sigma}_\theta(\boldsymbol{x}_t, t) = \exp(\boldsymbol{v}\log\beta_t + (1-\boldsymbol{v}\log\tilde{\beta}_t))$. Para integrar $\boldsymbol{\Sigma}_\theta(\boldsymbol{x}_t, t)$ en la función de pérdida, modifican la función original por $L_{\text{hibrida}} = L_{\text{simple}} + \lambda L_{\text{VLB}}$, con $\lambda = 0{,}001$, utilizando $\lambda L_{\text{VLB}}$ para aprender la varianza pero no la media debido a que comprueban experimentalmente que $\lambda L_{\text{VLB}}$ es difícil de optimizar debido a que tiene gradientes muy variables. Dadas estas circunstancia, para facilitar el proceso de aprendizaje proponen el uso de una media movil ponderada.

En [9] se utiliza un nuevo tipo de parametrización utilizando la razón de señal a ruido (SNR) y se muestra como optimizar la reversión minimizando un objetivo variacional.

Mediante la minimización de un objetivo variacional, otros trabajos entrenan $\sigma_t^2$ al mismo tiempo que se optimiza el modelo de difusión [8] y otros después [10].

### 3.7 Conexión con los VAEs

Los modelos de difusión pueden ser considerados como una forma particular de VAEs jerárquicos. Las diferencias esenciales residen en el hecho que el codificador está prefijado, las dimensiones del espacio latente son las mimas que las de los datos originales, el modelo de eliminación de ruido es el mismo para todos los pasos y que el modelo es entrenado con alguna modificación de los pesos que forman el límite variacional.

### 3.8 Resumen

En esta sección hemos introducido los modelos probabilísticos de eliminación del ruido por difusión. El modelo se entrena tomando muestras de un proceso de difusión directa a través del entrenamiento de un modelo de reversión capaz de predecir el ruido. Los puntos claves del diseño residen en la arquitectura de red escogida, la ponderación de los objetivos de optimización y la elección de los parámetros de difusión que controlan el proceso (como la distribución del ruido de generación). En [11] se puede encontrar un compendio de las decisiones de diseño relevantes a la hora de diseñar un modelo de difusión.

## 4 Generación basada en puntuación mediante el uso de ecuaciones diferenciales

### 4.1 Recordatorio de ecuaciones diferenciales

Las ecuaciones diferenciales ordinarias (ODE) tienen la forma siguiente:

$$\frac{d\boldsymbol{x}}{dt} = \boldsymbol{f}(\boldsymbol{x}, t)$$

La solución analítica de este tipo de ecuaciones tiene la forma:

$$\boldsymbol{x}(t) = \boldsymbol{x}(0) + \int_0^t \boldsymbol{f}(\boldsymbol{x}, \tau)d\tau$$

La solución numérica se puede conseguir iterativamente mediante:

$$\boldsymbol{x}(t + \Delta t) \approx \boldsymbol{x}(t) + \boldsymbol{f}(\boldsymbol{x}(t), t)\Delta t$$

Por otro lado, las ecuaciones diferenciales estocásticas (SDE) toman la forma siguiente:

Las ecuaciones diferenciales ordinarias (ODE) tienen la forma siguiente:

$$\frac{d\boldsymbol{x}}{dt} = \boldsymbol{f}(\boldsymbol{x}, t) + \sigma(\boldsymbol{x}, t)\boldsymbol{\omega}_t$$

donde a $\boldsymbol{f}(\boldsymbol{x}, t)$ se le denomina coeficiente de deriva y a $\boldsymbol{\sigma}(\boldsymbol{x}, t)$, coeficiente de difusión.

La solución numérica en este caso toma la forma siguiente:

$$\boldsymbol{x}(t + \Delta t) \approx \boldsymbol{x}(t) + \boldsymbol{f}(\boldsymbol{x}(t), t)\Delta t + \sigma(\boldsymbol{x}(t), t)\sqrt{\Delta t}\mathcal{N}(\boldsymbol{0}, \boldsymbol{I})$$

### 4.2 El proceso de difusión directa como una SDE

Consideremos de nuevo el proceso de difusión directa definido en la sección anterior:

$$q(\boldsymbol{x}_t, \boldsymbol{x}_{t-1}) = \mathcal{N}(\boldsymbol{x}_t; \sqrt{1 - \beta_t}\boldsymbol{x}_{t-1}, \beta_t\boldsymbol{I})$$

Si consideramos pasos infinitesimales podemos expresar a $\boldsymbol{x}_t$ como:

$$\begin{aligned} \boldsymbol{x}_t &= \sqrt{1 - \beta_t}\boldsymbol{x}_{t-1} + \sqrt{\beta_t}\mathcal{N}(\boldsymbol{0}, \boldsymbol{I}) \\ &= \sqrt{1 - \beta(t)\Delta t}\boldsymbol{x}_{t-1} + \sqrt{\beta(t)\Delta t}\mathcal{N}(\boldsymbol{0}, \boldsymbol{I}) \\ \text{donde}\, \beta_t = \beta(t)\Delta t & \end{aligned}$$

Aplicando la expansion de Taylor obtenemos:

$$\boldsymbol{x}_t \quad \approx \boldsymbol{x}_{t-1} - \tfrac{\beta(t)\Delta t}{2}\boldsymbol{x}_{t-1} + \sqrt{\beta(t)\Delta t}\mathcal{N}(\boldsymbol{0}, \boldsymbol{I})$$

De la que, llevándola al límite, podemos derivar la Ecuación diferencial estocástica (SDE) que describe la difusión en el límite infinitesimal:

$$d\boldsymbol{x}_t \quad = -\tfrac{1}{2}\beta(t)\boldsymbol{x}_t dt + \sqrt{\beta(t)}d\boldsymbol{\omega}_t$$

Atendiendo a la nomenclatura de ecuaciones diferenciales, el término $-\frac{1}{2}\beta(t)\boldsymbol{x}_t dt$ se correspondería con el término de deriva y el término $\sqrt{\beta(t)}d\omega_t$ con el término de difusión, que es el que inyecta el ruido.

Una ecuación más general para el SDE utilizada en modelos generativos de difusión sería la siguiente:

$$d\boldsymbol{x}_t = f(t)\boldsymbol{x}_t dt + g(t)d\boldsymbol{\omega}_t$$

## 4.3 El proceso de difusión inverso mediante SDEs

Visto el proceso directo, ¿Cómo se lleva a cabo el proceso inverso? La ecuación para llevar a cabo el proceso inverso generativo de difusión, introducida en [12] y propuesta para el proceso de generación de imágenes en [13] es la siguiente:

$$d\boldsymbol{x}_t = \Big[ -\frac{1}{2}\beta(t)\boldsymbol{x}_t - \beta(t)\nabla_{\boldsymbol{x}_t} \log q_t(\boldsymbol{x}_t)\Big] dt + \sqrt{\beta(t)}d\bar{\boldsymbol{\omega}}_t$$

donde $\nabla_{\boldsymbol{x}_t} \log q_t(\boldsymbol{x}_t)$ es la función de puntuación, $\sqrt{\beta(t)}d\bar{\boldsymbol{\omega}}_t$ el término de difusión. $-\frac{1}{2}\beta(t)\boldsymbol{x}_t - \beta(t)\nabla_{\boldsymbol{x}_t} \log q_t(\boldsymbol{x}_t)$ representa al término de deriva.

## 4.4 Función de puntuación

Un aspecto importante es definir la manera de determinar la función de puntuación. Una estrategia podría consistir en intentar modelizar dicha función mediante una regresión:

$$\min_{\theta} \mathbb{E}_{t\sim\mathcal{U}(0,T)}\mathbb{E}_{\boldsymbol{x}_t\sim q_t(\boldsymbol{x}_t)}||s_{\boldsymbol{\theta}}(\boldsymbol{x}_t, t) - \nabla_{\boldsymbol{x}_t} \log q_t(\boldsymbol{x}_t)||_2^2$$

Donde $t$ es el paso temporal de difusión, $\boldsymbol{x}_t$ los datos difundidos, $s_{\boldsymbol{\theta}}$ la red neuronal que pretendería modelizar a la función de puntuación y $\nabla_{\boldsymbol{x}_t} \log q_t(\boldsymbol{x}_t)$ la puntuación marginal de los datos difundidos [14]. El problema es que $\nabla_{\boldsymbol{x}_t} \log q_t(\boldsymbol{x}_t)$ resulta intratable.

En vez de utilizar el marginal, se puede realizar la difusión de puntos individuales aprovechando el hecho de que $q_t(\boldsymbol{x}_t|\boldsymbol{x}_0)$ es tratable. De esta manera podemos expresar la función objetivo del siguiente modo:

$$\min_{\theta} \mathbb{E}_{t\sim\mathcal{U}(0,T)}\mathbb{E}_{\boldsymbol{x}_0\sim q_0(\boldsymbol{x}_0)}\mathbb{E}_{\boldsymbol{x}_t\sim q_t(\boldsymbol{x}_t)|\boldsymbol{x}_0)}||s_{\boldsymbol{\theta}}(\boldsymbol{x}_t, t) - \nabla_{\boldsymbol{x}_t} \log q_t(\boldsymbol{x}_t|\boldsymbol{x}_0)||_2^2$$

Después de aplicadas las esperanzas resulta que, $s_{\boldsymbol{\theta}}(\boldsymbol{x}_t, t) \approx \nabla_{\boldsymbol{x}_t} \log q_t(\boldsymbol{x}_t)$ que es precisamente la función que pretendemos aproximar.

Para la SDE que preserva la varianza sabemos que:

$$\begin{gathered} d\boldsymbol{x}_t = -\tfrac{1}{2}\beta(t)\boldsymbol{x}_t dt + \sqrt{\beta(t)}d\boldsymbol{\omega}_t \\ q_t(\boldsymbol{x}_t|\boldsymbol{x}_0) = \mathcal{N}(\boldsymbol{x}_t; \gamma_t\boldsymbol{x}_0, \sigma_t^2\boldsymbol{I}) \\ \gamma_t = e^{-\frac{1}{2}\int_0^t \beta(s)ds} \\ \sigma_t^2 = 1 - e^{-\int_0^t \beta(s)ds} \end{gathered}$$

A modo de detalles de implementación, el muestreo se puede re-expresar del siguiente modo:

$$\begin{aligned} \text{Muestreo:} \quad & \boldsymbol{x}_t = \gamma_t\boldsymbol{x}_0 + \sigma_t\boldsymbol{\epsilon} \qquad \boldsymbol{\epsilon} \sim \mathcal{N}(\boldsymbol{0}, \boldsymbol{I}) \\ \text{Puntuación:} \quad & \nabla_{\boldsymbol{x}_t} \log q_t(\boldsymbol{x}_t|\boldsymbol{x}_0) = \nabla_{\boldsymbol{x}_t} \tfrac{(\boldsymbol{x}_t-\gamma_t\boldsymbol{x}_0)^2}{2\sigma_t^2} = -\tfrac{\boldsymbol{\epsilon}}{\sigma_t} \\ \text{Red neuronal modelo:} \quad & s_{\boldsymbol{\theta}}(\boldsymbol{x}_t, t) := \tfrac{\boldsymbol{\epsilon}_{\boldsymbol{\theta}}(\boldsymbol{x}_t, t)}{\sigma_t} \end{aligned}$$

Con lo que la función de optimización se puede expresar del siguiente modo:

$$\min_{\theta} \mathbb{E}_{t\sim\mathcal{U}(0,T)}\mathbb{E}_{\boldsymbol{x}_0\sim q_0(\boldsymbol{x}_0)}\mathbb{E}_{\boldsymbol{\epsilon}\sim\mathcal{N}(\mathbf{0},\boldsymbol{I})}\frac{1}{\sigma_t^2}||\boldsymbol{\epsilon}-\boldsymbol{\epsilon}_{\boldsymbol{\theta}}(\boldsymbol{x}_t,t)||_2^2$$

Donde el la red neuronal alternativa a optimizar, para predicción del ruido es $\boldsymbol{\epsilon}_{\boldsymbol{\theta}}(\boldsymbol{x}_t,t)$.

Una modificación adicional que se puede implementar es introducir una variable $\lambda(t)$ que permita modificar el peso de los distintos constituyentes de la función de pérdida; esto es, de las diferentes partes del proceso de difusión:

$$\min_{\theta} \mathbb{E}_{t\sim\mathcal{U}(0,T)}\mathbb{E}_{\boldsymbol{x}_0\sim q_0(\boldsymbol{x}_0)}\mathbb{E}_{\boldsymbol{\epsilon}\sim\mathcal{N}(\mathbf{0},\boldsymbol{I})}\frac{\lambda(t)}{\sigma_t^2}||\boldsymbol{\epsilon}-\boldsymbol{\epsilon}_{\boldsymbol{\theta}}(\boldsymbol{x}_t,t)||_2^2$$

Modificando su valor es posible optimizar para la obtención de buenos resultados perceptuales $\lambda(t)=\sigma_t^2$ o bien para la maximización de la verosimilitud $\lambda(t)=\beta(t)$.

Hemos presentado esta posibilidad pero es posible la utilización de modelos con parametrizaciones más sofisticadas [11].

Finalmente, en [15] se introducen una serie de modificaciones destinadas a reducir la varianza y mejorar la estabilidad numérica que se traducen en la siguiente propuesta de función de pérdida:

$$\min_{\theta} \mathbb{E}_{t\sim r(t)}\mathbb{E}_{\boldsymbol{x}_0\sim q_0(\boldsymbol{x}_0)}\mathbb{E}_{\boldsymbol{\epsilon}\sim\mathcal{N}(\mathbf{0},\boldsymbol{I})}\frac{1}{r(t)}\frac{\lambda(t)}{\sigma_t^2}||\boldsymbol{\epsilon}-\boldsymbol{\epsilon}_{\boldsymbol{\theta}}(\boldsymbol{x}_t,t)||_2^2$$

Donde $r(t)$ es una función de muestreo de la importancia. $r(t)\propto\frac{\lambda(t)}{\sigma_t^2}$. Para detalles concretos de implementación consultar la bibliografía.

## 5 Conclusiones

En este capítulo hemos brindado una introducción a los modelos de difusión, los cuales tienen la capacidad de revertir el proceso de difusión directa y recuperar la imagen original que ha sido degradada. Hemos explorado dos enfoques para modelar esta reversión: los modelos probabilísticos y los basados en ecuaciones diferenciales. No obstante, es importante destacar que el campo de los modelos de difusión es mucho más amplio y complejo que lo que se ha presentado en este artículo, y abarca muchos otros aspectos que no hemos podido cubrir en su totalidad. Este tema es uno de los más relevantes en el ámbito de los modelos generativos, y está en constante evolución y expansión. Para poder profundizar en este tema es necesario complementar esta información con las referencias y artículos futuros. El objetivo de esta introducción es sentar las bases para poderse adentrar en este complejo espacio de conocimiento de manera autónoma y continuar explorando las múltiples posibilidades que ofrece el uso de modelos de difusión en la generación de imágenes.

## Referencias